# Evaluating Las Vegas Algorithms — Pitfalls and Remedies

Holger H. Hoos and Thomas Stützle
Computer Science Department
Darmstadt University of Technology
D-62483 Darmstadt, Germany
{hoos,stuetzle}@informatik.tu-darmstadt.de

## Abstract

Stochastic search algorithms are among the most sucessful approaches for solving hard combinatorial problems. A large class of stochastic search approaches can be cast into the framework of Las Vegas Algorithms (LVAs). As the run-time behavior of LVAs is characterized by random variables, the detailed knowledge of run-time distributions provides important information for the analysis of these algorithms. In this paper we propose a novel methodology for evaluating the performance of LVAs, based on the identification of empirical run-time distributions. We exemplify our approach by applying it to Stochastic Local Search (SLS) algorithms for the satisfiability problem (SAT) in propositional logic. We point out pitfalls arising from the use of improper empirical methods and discuss the benefits of the proposed methodology for evaluating and comparing LVAs.

## 1 INTRODUCTION

Las Vegas algorithms are nondeterministic algorithms with the following properties: If a solution is returned, it is guaranteed to be correct, and the run-time is characterized by a random variable. Las Vegas algorithms are prominent not only in the field of Artificial Intelligence but also in other areas of computer science and Operations Research. In the recent years stochastic local search (SLS) algorithms such as Simulated Annealing, Tabu Search, and Evolutionary Algorithms have been found to be very successful for solving NP-hard problems from a broad range of domains. But also a number of systematic search methods, like some modern variants of the Davis Putnam algorithm for propositional satisfiability (SAT) problems, or backtracking-style algorithms for CSPs and graph coloring problems make use of non-deterministic decisions (like randomized tie-breaking rules) and can thus be characterized as Las Vegas algorithms.

Due to their non-deterministic nature, the behavior of Las Vegas algorithms is usually difficult to analyze. Even in the cases where theoretical results do exist, their practical applicability is often very limited, as in the case of Simulated Annealing, which is proven to converge towards an optimal solution under certain conditions which, however, cannot be met in practice. Given this situation, in most cases analyses of the run-time behavior of Las Vegas algorithms are based on empirical methodology. In a sense, despite dealing with completely specified algorithms which can be easily understood on a step-by-step execution basis, computer scientists are in the same situation as, say, an experimental physicist observing some non-deterministic quantum phenomenon.

The methods that have been applied for the analysis of Las Vegas algorithms in AI, however, are rather simplistic. Nevertheless, at the first glance, these methods seem to be admissible, especially since the results are usually quite consistent in a certain sense. In case of SLS algorithms for SAT, for instance, advanced algorithms like WSAT (Selman et al., 1994) usually outperform older algorithms (such as GSAT (Selman et al., 1992)) on a large number of problems from both randomized distributions and structured domains. The claims which are supported by empirical evidence are usually relatively simple (like "algorithm A outperforms algorithm B"), and the analytical methodology used is both easy to apply and powerful enough to get the desired results.

Or is it really? Recently, there has been some severe criticism regarding the empirical testing of algorithms (Hooker, 1994; Hooker, 1996; McGeoch, 1996). It has been pointed out that the empirical methodology that is used to evaluate and compare algorithms does not reflect the standards which have been established in other empirical sciences. Also, it was argued that the empirical analysis of algorithms should not remain at the stage of collecting data, but should rather attempt to formulate hypotheses based on this data which, in turn, can be experimentally verified or refuted. Up to now, most work dealing with the empirical analysis of Las Vegas algorithms in AI has not lived up to these demands. Instead, recent studies still use basically the same methods that have been around for years, often investing tremendous computational effort in doing large scale experiments (Parkes and Walser, 1996) in order to ensure that the basic descriptive statistics are sufficiently stable. At the same time more fundamental issues, such as the question whether the particular type of statistical analysis that is done (usually estimating means and standard deviations) is adequate for the type of evaluation that is intended, are often neglected or not ad-dressed at all.

In this work, we approach the issue of empirical methodology for evaluating Las Vegas algorithms for decision problems, like SAT or CSP, in the following way. After discussing fundamental properties of Las Vegas algorithms and different application scenarios, we present a novel kind of analysis based on estimating the run-time distributions for single instances. We motivate why this method is superior to established procedures while generally not causing additional computational overhead. We then point out some pitfalls of improperly chosen empirical methodology, and show how our approach avoids these while additionally offering a number of benefits regarding the analysis of individual algorithms, comparative studies, the optimization of critical parameters, and parallelization.

## 2  LAS VEGAS ALGORITHMS AND APPLICATION SCENARIOS

An algorithm $A$ is a *Las Vegas algorithm* for problem class $\Pi$, if (i) whenever for a given problem instance $\pi \in \Pi$ it returns a solution $s$, $s$ is guaranteed to be a valid solution of $\pi$, and (ii) on each given instance $\pi$, the run-time of $A$ is a random variable $RT_{A,\pi}$. According to this definition, Las Vegas algorithms are always correct, while they are not necessarily complete. Since completeness is an important theoretical concept for the study of algorithms, we classify Las Vegas algorithms into the following three categories:

- *complete Las Vegas algorithms* can be guaranteed to solve each soluble problem instance within run-time $t_{max}$, where $t_{max}$ is an instance-dependent constant. Let $P(RT_{A,\pi} \leq t)$ denote the probability that $A$ finds a solution for a soluble instance $\pi$ in time $\leq t$, then $A$ is complete exactly if for each $\pi$ there exists some $t_{max}$ such that $P(RT_{A,\pi} \leq t_{max}) = 1$.

- *approximately complete Las Vegas algorithms* solve each soluble problem instance with a probability converging to 1 as the run-time approaches $\infty$. Thus, $A$ is approximately complete, if for each soluble instance $\pi$, $\lim_{t \to \infty} P(RT_{A,\pi} \leq t) = 1$.

- *essentially incomplete Las Vegas algorithms* are Las Vegas algorithms which are not approximately complete (and therefore also not complete).

Examples for complete Las Vegas algorithms in AI are randomized systematic search methods like modern Davis Putnam variants (Crawford and Auton, 1996). Many of the most prominent stochastic local search methods, like Simulated Annealing or GSAT with Random Walk, are approximately complete, while others, such as basic GSAT (without restart) and most variants of Tabu Search are essentially incomplete.

In literature, approximate completeness is often referred to as *convergence*. Convergence results are established for a number of SLS algorithms, such as Simulated Annealing. Approximate completeness can be enforced for most SLS algorithms by providing a restart mechanism, as can be found in GSAT (Selman et al., 1992). However, both forms of approximate completeness are mainly of theoretical interest, since the time limits for finding solutions are usually far too large to be of practical use.

### APPLICATION SCENARIOS

Before even starting to evaluate any algorithm, it is crucial to find the right evaluation criteria. Especially for Las Vegas algorithms there are fundamentally different criteria for evaluation, depending on the characteristics of the environment they are supposed to work in. Thus, we classify possible application scenarios in the following way:

**Type 1:** There are no time limits, i.e., we can afford to run the algorithm as long as it needs to find a solution. Basically, this scenario is given whenever the computations are done off-line or in a non-realtime environment, where it does not really matter how long we need to find a solution.

**Type 2:** There is a time limit $t_{max}$ for finding a solution. In real-time applications, like robotic control or dynamic scheduling, $t_{max}$ can be very small.

**Type 3:** The usefulness or utility of a solution depends on the time needed to find it. Formally, if utilities are represented as values in $[0, 1]$, we can characterize these scenarios by specifying a utility function $U : R \mapsto [0, 1]$, where $U(t)$ is the utility of finding a solution after time $t$. As can be easily seen, types 1 and 2 are special cases of type 3.

Obviously, different criteria are required for evaluating the performance of Las Vegas algorithms in these scenarios. While in the case of no time limits being given (type 1), the mean run-time might suffice to roughly characterize the run-time behavior, in real-time situations (type 2) it is basically meaningless. An adequate criterion for a type 2 situation with time-limit $t_{max}$ is $P(RT \leq t_{max})$, the probability of finding a solution within the given time-limit. For type 3, the most general scenario, the run-time behavior can only be adequately characterized by the run-time distribution function $rtd : R \mapsto [0, 1]$ defined as $rtd(t) = P(RT \leq t)$ or some approximation of it. The run-time distribution (RTD), however, completely and uniquely characterizes the run-time behavior of a Las Vegas algorithm. Given this information, other criteria, like the mean run-time, its standard deviation, median, percentiles, or success-probabilities $P(RT \leq t')$ for arbitrary time-limits $t'$ can be easily obtained.

## 3  OUR EMPIRICAL METHOD

To answer the questions that arise from the different application scenarios discussed in the previous section, it is important to have knowledge of the actual run-time distributions (RTD) of Las Vegas algorithms. The run-time is a random variable and we can get knowledge on its distribution by empirically taking samples of the random variable by simply running the algorithm several times. Based on the sample, assumptions on the type of distribution function can be made. These assumptions can be validated by statistical hypothesis tests and in case the assumptions cannot be backed up by the sample data, incorrect assumptions are identified and rejected. As the actual run-time distribution of an Las Vegas algorithm depends on the problem instance, it should be clear that RTDs should be estimated on single problem instances. Yet, this does not limit the type of conclusions that may be drawn as general conjectures on the type of run-time distributions over a whole problem class can be formed and tested. We will illustrate this point in more detail in Section 5.1.

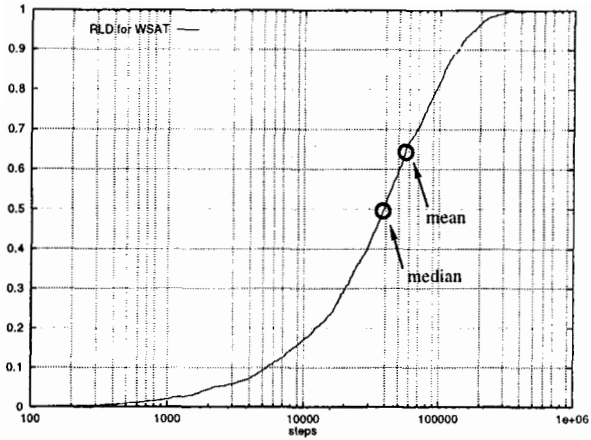

Figure 1: RLD for WSAT on a hard Random-3-SAT instance for optimal parameter settings; median vs. mean.

Instead of actually measuring run-*time* distributions in terms of CPU-time, it is often preferable to use representative operation counts as a more machine independent measure of an algorithm's performance. Using an appropriate cost model of the algorithm's operations, operation counts can easily be converted into run-times, facilitating comparisons of algorithms across architectures. Thus, instead of run-*time* distributions we get run-*length* distributions (RLDs). For example, an appropriate operation count for local search algorithms for SAT is the number of local search steps. In the following we will use run-time distributions and run-length distributions interchangeably as long as they can be converted one into the other.

To actually measure RLDs, one has to take into account that most algorithms have some cutoff parameter like maximum number of iterations, maximum time limit, or others. Practically, we measure empirical RLDs by running the respective Las Vegas algorithm for $n$ times on a given problem instance up to some (very high) cutoff value[1] and recording for each successful run the number of steps required to find a solution. The empirical run-length distribution is the cumulative distribution associated with the observations. More formally, let $rl(j)$ denote the run length for the $j$th successful run, then the cumulative empirical RLD is defined by $\widehat{P}(rl \leq i) = |\{j | rl(j) \leq i\}|/n$. Note, that obtaining run-length distributions for single instances does not involve a significantly higher computational effort than to get a stable estimate for the mean performance of an algorithm.

To give an example of an actually occuring RLD, we run a state-of-the-art local search algorithm (WSAT (Selman et al., 1994)) on a hard Random-3-SAT instance for optimal walk-parameter settings and present the RLD in Figure 1. The x-axis represents the computational effort as the number of local search steps, the y-axis gives the *empirical* success probability. One may note that the shape of the RLD is that of an exponential distribution $ed[m]$, with distribution function[2] $F(x) = 1 - 2^{-x/m}$. Actually, using a $\chi^2$-test, the hypothesis that the RLD corresponds to an exponential distribution passed the test. We will discuss potential benefits of our method in more detail in Section 5.

## 4  PITFALLS OF INADEQUATE METHODOLOGY

### 4.1  SUPERFICIAL ANALYSIS OF RUN-TIME BEHAVIOR

A well-established method for evaluating the run-time behavior of Las Vegas algorithms is to measure average run-times on single instances in order to obtain an estimate of the mean run-time. Practically, this is done by executing the algorithm $n$ times on a given problem instance with cutoff time $t_{max}$. If $k$ of these runs are successful and $rt_i$ is the run-time of the $i$th successful run, the mean run-time is estimated by averaging over the successful runs and accounting for the expected number of runs required to find a solution:

$$\widehat{E}(RT) = \frac{1}{k}\sum_{i=1}^{k} rt_i + (n-k)/k \cdot t_{max} \quad (1)$$

One problem with this method is that the mean alone gives only a very unprecise impression of the run-time behavior, even if additionally the standard deviation (or variance) for the run-time of the successful runs is reported. Consider the design of an algorithm for a type 2 application scenario and the specific question of estimating the cutoff time $t_{max}$ for solving a given problem instance with a probability $p$. If only the mean run-time $E(RT)$ is known, the best estimate we can obtain is given by the Markov inequality (Rohatgi, 1976) $P(RT \geq t) \leq E(RT)/t$:

$$t_{max} = E(RT)/(1-p) \quad (2)$$

If the standard deviation $\sigma(RT)$ is known and finite, using the Tchebichev inequality $P(|RT - E(RT)| \geq \epsilon) \leq \sigma^2(RT)/\epsilon^2$ we obtain a better estimate:

$$t_{max} = \sigma(RT)/\sqrt{1-p} + E(RT) \quad (3)$$

If, however, we know that the run-time of a given Las Vegas algorithm is exponentially distributed[3], we get a much more accurate estimate. In the following example, we see the drastic differences between these three estimates. For a given Las Vegas algorithm applied to some problem the mean run-time and the standard deviation are 100 seconds each, a situation which is not untypical, e.g., for stochastic local search algorithms for SAT. We want to determine the run-time $t'$ required for obtaining a solution probability of 0.99. Without any additional knowledge on the run-time distribution we get an estimate of 1100 sec (using the Tchebichev inequality). If even the standard deviation is unknown, we can only use the Markov inequality and estimate $t'$ as 10000 sec! But assuming that the run-time is exponentially distributed, we get an estimate of 460 sec. This illustrates that as long as the type of RTD is not known *a priori*, analyzing only means and standard deviations is a very wasteful use of empirical data.

---
[1] Optimal cutoff settings may then be determined a posteriori using the empirical run-time distribution, see Sec. 5.2.

[2] In statistical literature, typically the exponential distribution is presented with respect to base $e$. With base 2, instead, the advantage is that parameter $m$ corresponds to the median of the distribution.

[3] As we will discuss later, assuming an exponential RTD is quite realistic, since we found that this can be observed for a number of modern stochastic local search algorithms on various problem classes.

Another problem, especially in recent literature on stochastic local search, lies in the tacit assumption that several parameters of the considered algorithms can be studied independently. In specific cases, it is known that this assumption does not hold (Hoos and Stützle, 1996; Steinmann et al., 1997). For the evaluation of Las Vegas algorithms in general, it is crucial to be aware of possible parameter dependencies, especially those involving the cutoff time $t_{max}$ which plays an important role in type 2 and 3 application scenarios.

In Fig. 2, we show the RLDs of two different Las Vegas algorithms LVA 1 and LVA 2 for the same problem instance. As can be easily seen, LVA 1 is essentially incomplete with an asymptotic solution probability approaching ca. 0.09, while LVA 2 is approximately complete. Now, note the crossing of the two RLDs at ca. 120 steps. For smaller cutoff times, LVA 1 achieves considerably higher solution probabilities, while for greater cutoff times, LVA 2 shows increasingly superior performance. Actually, using the optimal cutoff time of ca. 57 steps in connection with restart, the exponential run-time distribution marked "ed[744]" can be obtained, which realizes a speedup of ca. 24% compared to LVA 2. Thus, the performance of LVA 1 is not only superior to that of LVA 2 for small cutoff times, but based on the RTDs, it is possible to modify algorithm LVA 1 such that its overall performance dominates that of LVA 2, see also Sec. 5.

As a consequence of these observations, basing the comparison of Las Vegas algorithms on expected run times is in the best case unprecise, in the worst case it leads to erroneous conclusions. The latter case occurs, when the two corresponding RTDs have at least one intersection. Then, obviously, the outcome of comparing the two algorithms depends entirely on the cutoff time $t_{max}$ which was chosen for the experiment.

### 4.2 INHOMOGENEOUS TEST SETS

Often, Las Vegas algorithms are tested on sets of randomly generated problem instances. This method is particularly popular for problem classes, for which phase transition phenomena have been observed, such as Random-3-SAT (Crawford and Auton, 1996) or Random-CSP (Smith and

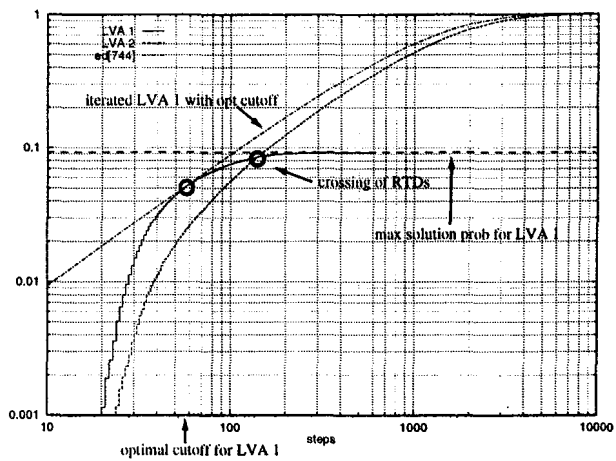

Figure 2: RLDs for two different Las Vegas algorithms on the same problem instance. Note the crossover of the two RLDs.

Dyer, 1996), because instances from the phase-transition region have been found to be particularly hard. Practically, the test sets for SLS algorithms are usually obtained by generating a number of sample instances from the phase transition area and filtering out unsolvable instances using a complete algorithm. The evaluation of Las Vegas algorithms on such a test set is done by evaluating a number of runs on each instance. Usually, the final performance measure is obtained by averaging over all instances from the test set.

This last step, however, is potentially extremely problematic. Since the run-time behavior on each single instance is characterized by a RTD (as discussed above), averaging over the test set is equivalent to averaging over these RTDs. Because in general, averaging over a number of distributions yields a distribution of a different type, the practice of averaging over RTDs (and thus the averaging over test sets) is quite prone to producing observations, which do not reflect the behavior of the algorithm on individual instances, but rather a side-effect of this method of evaluation. We exemplify this for Random-3-SAT near the phase-transition, a problem class which has been used in many studies of stochastic local SAT procedures, such as GSAT or GSAT with random walk (GWSAT) (Selman et al., 1992; Gent and Walsh, 1995). Our own experimental studies have shown, that for GWSAT with optimal walk parameter (as well as for a number of other stochastic local search algorithms, such as WSAT (Selman et al., 1994) or NOVELTY (McAllester et al., 1997)), the RLDs on single instances from this problem distribution can be reasonably well approximated by exponential distributions (Hoos and Stützle, 1998). By measuring the median run-lengths for each problem instance from a typical randomly generated and filtered test set, we obtain a distribution of the median hardness of the problems as shown in Fig. 3. Since the median run-lengths were determined using 1000 runs per instance, they are very stable which leaves the random selection of the instances as the main source of noise in the measured distribution.

Since the RLDs for single instances are basically exponen-tial distributions, optimal cutoff times cannot be observed for individual instances. The reason for this is the fact that for exponential run-time distributions, the probability of finding a solution within $k$ steps of the algorithm is independent of the number of steps performed before. (This is discussed in more detail in Sec. 5.) Since the class of exponential distributions is *not* closed under averaging, the combined RLD for all instances in the test set, obtained by averaging over the individual RLDs, is *not* exponentially distributed. But for this combined distribution, shown in Fig. 4, obviously an optimal cutoff time exists, which is obtained by finding the minimal value $m^*$ for which $ed[m^*]$ and the average RLD have at least one common point.

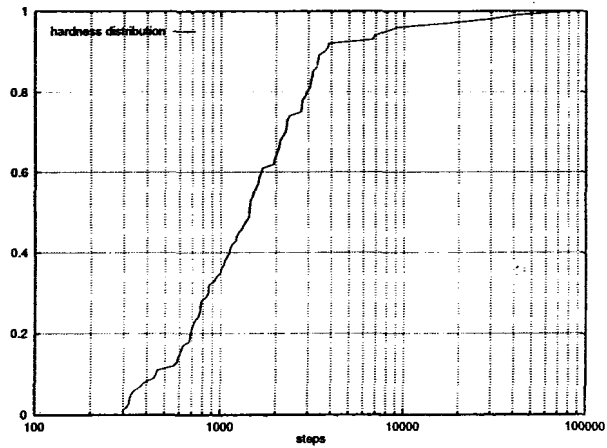

Figure 3: Distribution of median run-length for WSAT (noise=55/100) over a test set of 100 satisfiable Random-3-SAT formuale (100 variables, 430 clauses). Note the huge variance, especially for the hardest 10% of the test set.

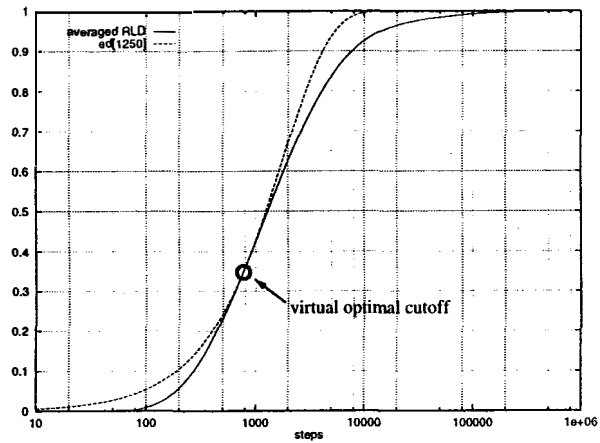

Figure 4: Average RLD for WSAT on the same test set. Note the virtual optimal cutoff at ca. 750 steps, for details see text.

Thus, while the averaged RLD suggests the existence of an overall optimal cutoff time, actually for each single instance, an optimal cutoff time does not exist. Although this might seem a bit paradoxical, this observation can be easily explained: When averaging over the RLDs, we don't distinguish between the probability of solving different instances. Using the "optimal" cutoff inferred from the average RLD then simply means that solving some easier instances with a sufficiently high probability compensates for the very small solution probability for the harder instances going along with using this cutoff. Thus, solving easier instances gets priority over solving harder instances. Under this interpretation, the "optimal" cutoff can be considered meaningful. Assuming, however, that in practice the goal in testing Las Vegas algorithms on test sets sampled from random problem distributions is to get a realistic impression of the overall performance, *including hard instances*, the "optimal" cutoff inferred from the averaged RLD is substantially misleading.

The above discussion shows, that by averaging over test sets, generally in the best case one consciously observes a bias for solving certain problems, the practical use of which seems rather questionable. But far more likely, being not aware of these phenomena, the observations thus obtained are misinterpreted and lead to erroneous conclusions. One could, however, imagine a situation in which averaging over test sets is not that critical. This would be given if the test sets are very homogeneous in the sense that the RTDs for each single instance are roughly identical. Unfortunately, this sort of randomized problem distributions seems to be very rare in practice. Certainly, Random-3-SAT is not homogeneous in this sense, and at least the authors are currently not aware of any sufficiently complex homogenous randomized problem distribution for SAT or CSP.[4]

Generally, a fundamental problem with averaging over random problem distributions is the mixing of two different sources of randomness in the evaluation of algorithms: the nondeterministic nature of the algorithm itself, and the random selection of problem instances. Assuming that in the analysis of Las Vegas algorithms one is mostly interested in the properties of the algorithm, at least a very good knowledge of the problem distribution is required for separating the influence of these inherently different types of randomness.

## 5   BENEFITS OF OUR METHOD

### 5.1   CHARACTERIZING RTDs

In this section we demonstrate our empirical methodology for testing Las Vegas algorithms with stochastic local search as one example application and illustrate how interesting observations can be made using our approach. For this purpose we analyze the run-time behavior of GSAT with random walk (GWSAT) on a hard Random-3-SAT instance from the phase transition region using various walk-probability settings, including the optimal one. The walk-probability $wp$ is, besides the cutoff value, the most important parameter of this algorithm. The algorithm was run 200 times for different settings of $wp$. Based on the em-pirical distribution for the optimal noise-parameter setting $wp_{opt}$ we conjectured that the run-time distribution can be well approximated by an exponential distribution. With this hypothesis the parameter $m$ can be estimated and the goodness-of-fit of the empirical distribution can be tested. In our concrete example this hypothesis passed the $\chi^2$-test.

---

[4]There is, however, some indication, that certain randomized classes of SAT encoded problems from other domains, such as compactly encoded subclasses of the Hamilton Circuit Problem (Hoos, 1996), are significantly more homogeneous than Random-3-SAT.

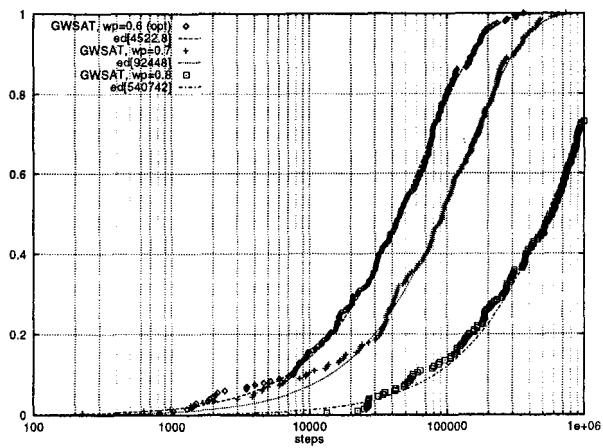

Figure 5: Run-time distribution for GSAT with random walk on a hard random 3-SAT instance for optimal and higher-than-optimal walk-parameter settings.

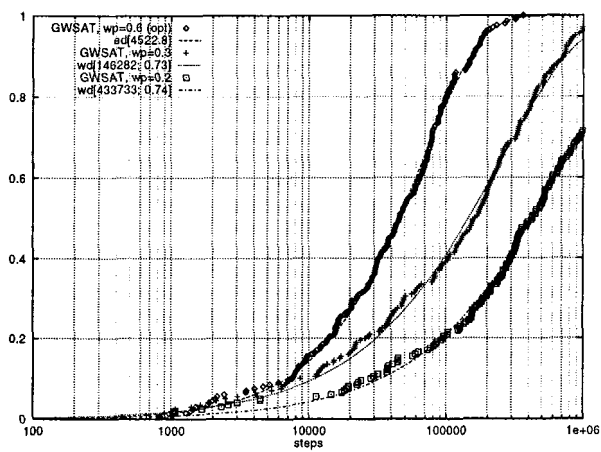

Figure 6: Run-time distribution for GSAT with random walk on a hard random 3-SAT instance for optimal and lower-than-optimal walk-parameter settings.

For larger than optimal noise-parameter settings $wp > wp_{opt}$ we could verify by the $\chi^2$-test that the RLDs are exponentially distributed. Yet, the RTDs are shifted to the right by a constant factor, if using a log-scale on the x-axis as in Figure 5. This means that to guarantee a desired success probability, if using $wp > wp_{opt}$ the required number of steps is by a constant factor higher than for $wp = wp_{opt}$. A completely different behavior can be observed for $wp < wp_{opt}$ (see Figure 6). In this case, the run-time distributions cannot be approximated by exponential distributions, but instead we found a good approximation using Weibull distributions $wd[m, \alpha]$ with $F(x) = 1 - 2^{-(x/m)^{\alpha}}$, $\alpha < 1$. This corresponds to the fact that the higher the desired solution probability, the larger will be the loss of performance

w.r.t. the optimal noise setting. Additionally, for instances which are easy to solve we could observe systematic deviations from the distribution assumptions in the lower part. These deviations may be explained by the initial hill-climb phase (Gent and Walsh, 1993), as, intuitively, it needs some time for the algorithm to reach a position in the search space for which there is a realistically high chance of finding a solution.

Our proposed way of measuring and analyzing RTDs shows considerable benefits. The statistical distributions of the RTDs can be identified and the goodness-of-fit can be tested using standard statistical tests. Also, based on such a methodology an experimental approach of testing Las Vegas algorithms along the lines proposed by Hooker (Hooker, 1996) can be undertaken. For example, based on the above discussion one such hypothesis is that for optimal noise parameter settings for GWSAT, its run-time behavior for the solution of single, hard instances from the crossover region of Random-3-SAT can be well described by exponential distributions. This hypothesis can be tested by running a series of experiments on such instances. Doing so, our empirical evidence confirms this conjecture. Similar results have been established for other algorithms, and not only applied to Random-3-SAT but also on SAT-encoded problems from other domains (Hoos and Stützle, 1998). Thus, by studying run-time distributions on single instances, hypotheses on the algorithm's behavior on a problem class like random 3-SAT can be formed and tested experimentally. It is important to note that limiting the analysis of RTDs to single instances does not impose limitations on obtaining and verifying conjectures on the behavior of algorithms on whole problem classes. Furthermore, by measuring RLDs (RTDs) we could observe a qualitatively different behavior of GWSAT, depending on whether lower than optimal or larger than optimal walk-parameter settings are chosen. For $wp > wp_{opt}$, the run-time distribution can still be identified as an exponential distribution, i.e., the type of distribution does not change. On the other hand, if $wp < wp_{opt}$ the type of distribution changes. Based on these observations further fundamental hypotheses (theories) describing the behavior of this class of algorithms may be formed and experimentally validated.

## 5.2 COMPARING AND IMPROVING ALGORITHMS

Important aspects like the comparison and design of algorithms can be addressed by the study of RLDs. Consider the situation presented in Figure 7 in which RLDs for two specific SLS algorithms (LVA A and LVA B) are plotted. LVA B is approximately complete and its performance monotonically improves with increasing run lengths (as can be seen from the decreasing distance between the RLD and the projected optimal exponential distribution $ed[1800]$). LVA A is essentially incomplete, the success probability converges to ca. 0.08. For very short runs we also observe "convergent behavior": Apparently for both algorithms there exists a minimal (sample size dependent) number of steps (marked $s_1$ and $s_2$ on the x-axis) below that the probability to find a solution is negligible. Interestingly, both curves cross in one specific point at ca. 420 steps; i.e., without using restarts LVA A gives a higher solution probability than LVA B for shorter runs, whereas LVA B is more effective for longer runs. Yet, for LVA A an optimal cutoff value of ca. 170 steps exists. Thus, repeated execution of LVA A for a reasonably well chosen cutoff time, after which the algorithm is restarted, gives a much higher solution probability as actually observed when running the algorithm for a long time. In fact, if the optimal cutoff parameter is chosen, one more point of interest is at ca. 500 steps (marked $s_3$ on the x-axis): For a lower number of steps as $s_3$, using independent runs of LVA A with optimal cutoff value one improves upon the performance of LVA B, while past $s_3$ LVA B is strictly superior to LVA A. Consequently, an anytime combination of the two algorithms should select LVA A with optimal cutoff up to $s_3$ steps, and then switch to LVA B.

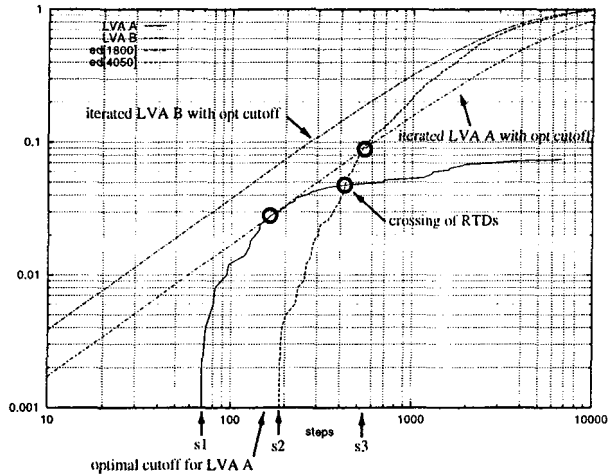

Figure 7: RLDs for two SLS algorithms (LVA A and LVA B) for the propositional satisfiability problem on a hard Random-3-SAT instance (the two RLDs cross over at ca. 420 steps), see text for a detailed discussion.

This example illustrates two important points. One concerns the comparison of algorithms. Generally, a Las Vegas algorithm dominates another one, if it consistently gives a higher solution probability. More formally, LVA 1 dominates LVA 2 if for optimal parameter settings $\forall t : P(RT_1 \leq t) \geq P(RT_2 \leq t)$ and $\exists t : P(RT_1 \leq t) > P(RT_2 \leq t)$. In case the RTDs cross over, the comparison of algorithms is substantially more difficult. Then, as detailed in the example above, often it occurs that one algorithm is preferable for lower time limits while the other may be better for long run times. In such a case an anytime combination of both algorithm can enhance overall performance.

A further, even more important point concerns the steepness of the run-time distributions. It is well known that for exponentially distributed run-times, due to the properties of the exponential distribution (Rohatgi, 1976), we get the same solution probability running an algorithm once for time $t$ or $p$ times for time $t/p$. If from some point on the run-time distribution of an algorithm is steeper than an exponential distribution, the probability of finding a solution relatively increases when running the algorithm for a longer time. In such a case it would be worse to restart the algorithm after some fixed cutoff value as can be seen for LVA B in Figure 7. On the other side, if the run-time distribution of an algorithm is less steep than an exponential distribution for increasing run-time, we can gain performance by restarting the algorithm as is the case for algorithm LVA A in Figure 7. This situation is given for many greedy local search algorithms like GSAT that easily get stuck in local minima. These algorithms usually have a run-time distribution that approaches a limiting success probability $< 1$, thus, they can gain a lot by restarts. In such a case optimal cutoff values can be identified using the method mentioned in Section 4. Due to the above mentioned property of the exponential distribution, for exponentially distributed run times an arbitrary cutoff time can be chosen. This last observation has also important consequences for parallel processing by multiple *independent* runs of an algorithm. Recall, that the speed-up in case of parallel processing is defined as $s = \frac{sequential\ time}{parallel\ time}$. For exponential RTDs, we would obtain the same probability of finding a solution on $p$ processors for time $t/p$, thus, resulting in an optimal speed-up. In case the RTDs are steeper than an exponential distribution, the resulting speed-up will be sub-optimal. On the other hand, if the run-time distribution is less steep than an exponential distribution, parallel processing even would yield a super-optimal speed-up when compared to the sequential version of the algorithm without restarts.

Summarizing our results, we argue that a detailed empirical analysis of algorithms using RTDs gives a more accurate picture of their behavior. Knowledge of this behavior can be very useful for formulating and testing hypotheses on an algorithm's behavior, improving algorithms, and devising efficient parallelizations.

## 6 RELATED WORK

Our work on the empirical evaluation of Las Vegas Algorithms is in part motivated by the theoretical work in (Alt et al., 1996; Luby et al., 1993). In (Alt et al., 1996), the authors discuss general policies for reducing the tail probability of Las Vegas algorithms. (Luby et al., 1993) discuss policies for selecting cutoff times for known and unknown run-time distributions.

Most work on the empirical analysis of algorithm for SAT and CSPs concentrates on analyzing cost distributions for *complete* search algorithms on randomly generated instances from random 3-SAT and binary CSPs. In (Frost et al., 1997) the cost distribution for randomly generated Random-3-SAT and binary CSPs from the phase transition region is approximated by continuous probability distributions. Yet, all these investigations concentrate on the cost distribution for a sample of instances from a fixed random distribution, not on the cost distribution of algorithms on single instances. A first approach investigating the cost distribution of a backtracking algorithm using the Brelaz heuristic on randomly generated 3-coloring problems on single instances is presented in (Hogg and Williams, 1994). Note, that the Brelaz heuristic breaks ties randomly, i.e., a backtracking algorithm using this heuristic is actually a Las

Vegas algorithm. The authors investigate the run-time distribution to make conjectures on the obtainable speed-up for parallel processing and find that the speed-up is strongly dependent on the cost distribution. Especially for multimodal cost distributions high speed-ups could be observed. A similar approach is taken in (Gomes and Selman, 1997), in which the authors intend to design algorithm portfolios using backtracking algorithms based on the Brelaz heuristic for a special kind of CSP. Finally, it should be noted that run-time distribution also have been observed occasionally in the Operations Research literature, like in (Taillard, 1991).

## 7   CONCLUSIONS

In this work, we introduced a novel approach for the empirical analysis of Las Vegas algorithms. Our method is based on measuring and analyzing run-time distributions (RTDs) for individual problem instances. Based on a classification of application scenarios for Las Vegas algorithms, we have shown that in general, only RTDs provide all the information required to adequately describe the behavior of the algorithm. Compared to the methodology which is commonly used for empirical analyses of Las Vegas algorithms in AI, our approach gives a considerably more detailed and realistic view of the behavior of these algorithms without requiring an additional overhead in data acquisition.

We identified and discussed two problems which are commonly arising in the context of inadequate empirical methodology: superficial analysis of the run-time behavior, and averaging over inhomogeneous test sets. As we have shown, our approach avoids the pitfalls arising from these. We further demonstrated, how our refined methodology can be used to obtain novel results in the characterization of the run-time behavior of some of the most popular stochastic local search algorithms in recent AI research.

In future work, we plan to extend our analysis of RTDs, trying to provide precise characterizations of the behavior of various state-of-the-art stochastic local search algorithms on a variety of problem classes. This includes the extension of our methodology for LVAs to optimization problems, like the Traveling Salesman Problem or scheduling problems.

Since to date, theoretical knowledge on the behavior of Las Vegas algorithms is very limited, an adequate empirical methodology is critical for investigating these algorithms. It is very likely that for the further improvement of these algorithms a deeper understanding of their behavior will be essential. We believe that in this context our improved and refined empirical methodology for analyzing the run-time behavior of Las Vegas algorithms in general, and SLS algorithms in particular, will prove to be very useful.

## References

Alt, H., Guibas, L., Mehlhorn, K., Karp, R., and Wigderson, A. (1996). A Method for Obtaining Randomized Algorithms with Small Tail Probabilities. *Algorithmica*, 16:543–547.

Crawford, J. and Auton, L. (1996). Experimental Results on the Crossover Point in Random 3SAT. *Artificial Intelligence*.

Frost, D., Rish, I., and Vila, L. (1997). Summarizing CSP Hardness with Continuous Probability Distributions. In *Proc. of AAAI'97*, pages 327–333.

Gent, I. and Walsh, T. (1993). An Empirical Analysis of Search in GSAT. *J. of Artificial Intelligence Research*, 1:47–59.

Gent, I. and Walsh, T. (1995). Unsatisfied Variables in Local Search. In *Hybrid Problems, Hybrid Solutions*, pages 73–85, Amsterdam. IOS Press.

Gomes, C. and Selman, B. (1997). Algorithm Portfolio Design: Theory vs. Practice. In *Proc. of UAI'97*, pages 190–197. Morgan Kaufmann Publishers.

Hogg, T. and Williams, C. (1994). Expected Gains from Parallelizing Constraint Solving for Hard Problems. In *Proc. of AAAI'94*, pages 331–336.

Hooker, J. (1994). Needed: An Empirical Science of Algorithms. *Operations Research*, 42(2):201–212.

Hooker, J. (1996). Testing Heuristics: We Have It All Wrong. *J. of Heuristics*, pages 33–42.

Hoos, H. and Stützle, T. (1996). A Characterization of GSAT's Performance on a Class of Hard Structured Problems. Technical Report AIDA–96–20, FG Intellektik, TU Darmstadt.

Hoos, H. and Stützle, T. (1998). Characterizing the Run-time Behavior of Stochastic Local Search. Technical Report AIDA–98–01, FG Intellektik, TU Darmstadt.

Hoos, H. H. (1996). Solving Hard Combinatorial Problems with GSAT — A Case Study. In *Proc. of KI'96*, volume 1137 of *LNAI*, pages 107–119. Springer Verlag.

Luby, M., Sinclair, A., and Zuckerman, D. (1993). Optimal Speedup of Las Vegas Algorithms. *Information Processing Letters*, 47:173–180.

McAllester, D., Selman, B., and Kautz, H. (1997). Evidence for Invariants in Local Search. In *Proc. of AAAI'97*, pages 321–326.

McGeoch, C. (1996). Towards an Experimental Method for Algorithm Simulation. *INFORMS J. on Computing*, 8(1):1–15.

Parkes, A. and Walser, J. (1996). Tuning Local Search for Satisfiability Testing. In *Proc. of AAAI'96*, pages 356–362.

Rohatgi, V. (1976). *An Introduction to Probability Theory and Mathematical Statistics*. John Wiley & Sons.

Selman, B., Kautz, H., and Cohen, B. (1994). Noise Strategies for Improving Local Search. In *Proc. of AAAI'94*, pages 337–343. MIT press.

Selman, B., Levesque, H., and Mitchell, D. (1992). A New Method for Solving Hard Satisfiability Problems. In *Proc. of AAAI'92*, pages 440–446. MIT press.

Smith, B. and Dyer, M. (1996). Locating the Phase Transition in Binary Constraint Satisfaction Problems. *Artificial Intelligence*, 81:155–181.

Steinmann, O., Strohmaier, A., and Stützle, T. (1997). Tabu Search vs. Random Walk. In *Proc. of KI'97*, volume 1303 of *LNAI*, pages 337–348. Springer Verlag.

Taillard, É. (1991). Robust Taboo Search for the Quadratic Assignment Problem. *Parallel Computing*, 17:443–455.